\documentclass{article} 
\usepackage{iclr2023_conference,times}

\usepackage{amsmath,amsfonts,bm}

\def\eqref#1{equation~\ref{#1}}

\def\1{\bm{1}}

\DeclareMathAlphabet{\mathsfit}{\encodingdefault}{\sfdefault}{m}{sl}
\SetMathAlphabet{\mathsfit}{bold}{\encodingdefault}{\sfdefault}{bx}{n}

\DeclareMathOperator*{\argmax}{arg\,max}
\DeclareMathOperator*{\argmin}{arg\,min}

\usepackage{hyperref}
\usepackage{url}

\usepackage[utf8]{inputenc} 
\usepackage[T1]{fontenc}    
\usepackage{hyperref}       
\usepackage{url}            
\usepackage{booktabs}       
\usepackage{amsfonts}       
\usepackage{nicefrac}       
\usepackage{microtype}      

\usepackage{times}
\usepackage{epsfig}
\usepackage{graphicx}
\usepackage{amsmath}
\usepackage{amssymb}

\usepackage{array}
\usepackage{multirow}
\usepackage{longtable}
\usepackage{tabularx}
\usepackage{subfigure}
\usepackage{float}
\usepackage{caption}
\usepackage{wrapfig}
\usepackage{makecell}
\usepackage{tabularx,tabulary}

\usepackage{xcolor}

\usepackage[linesnumbered,ruled]{algorithm2e}

\usepackage{algpseudocode}
\usepackage{arydshln}
\usepackage{pifont}
\usepackage{url}
\usepackage{enumitem}
\usepackage{booktabs}
\usepackage{leftidx}

\algnewcommand\algorithmicinput{\textbf{Input:}}
\algnewcommand\INPUT{\item[\algorithmicinput]}
\algnewcommand\algorithmicoutput{\textbf{Output:}}
\algnewcommand\OUTPUT{\item[\algorithmicoutput]}

\def\ie{\emph{i.e.}}

\title{Meta Knowledge Condensation for\\ Federated Learning}

\author{Ping Liu  \\
Center for Frontier AI Research\\
A*STAR\\
Singapore \\
\texttt{\{pino.pingliu\}@gmail.com} \\
\And
Xin Yu \\
Australian Artificial Intelligence Institute \\
University of Technology Sydney \\
Sydney, Australia \\
\texttt{\{xin.yu\}@uts.edu.au} \\
\AND
Joey Tianyi Zhou \\
Center for Frontier AI Research\\
A*STAR\\
Singapore \\
\texttt{\{joey.tianyi.zhou\}@gmail.com}
}

\begin{document}

\maketitle

\begin{abstract}
Existing federated learning paradigms usually extensively exchange distributed models at a central solver to achieve a more powerful model. 
However, this would incur severe communication burden between a server and multiple clients especially when data distributions are heterogeneous.
As a result, current federated learning methods often require a large number of communication rounds in training.
Unlike existing paradigms, we introduce an alternative perspective to significantly decrease the communication cost in federate learning.
In this work, we first introduce a meta knowledge representation method that extracts meta knowledge from distributed clients. 
The extracted meta knowledge encodes essential information that can be used to improve the current model.
As the training progresses, the contributions of training samples to a federated model also vary.
Thus, we introduce a dynamic weight assignment mechanism that enables samples to contribute adaptively to the current model update.
Then, informative meta knowledge from all active clients is sent to the server for model update.
Training a model on the combined meta knowledge without exposing original data among different clients
can significantly mitigate the heterogeneity issues.
Moreover, to further ameliorate data heterogeneity, we also exchange meta knowledge among clients as conditional initialization for local meta knowledge extraction.
Extensive experiments demonstrate the effectiveness and efficiency of our proposed method. Remarkably, our method outperforms the state-of-the-art by a large margin (from $74.07\%$ to $92.95\%$) on MNIST with a restricted communication budget (\textit{i.e.}, 10 rounds).
\end{abstract}

\section{Introduction}
\label{sec:introduction}

Most deep learning-based models are trained in a data-centralized manner. 
However, in some cases, data might be distributed among different clients and cannot be shared. 
To address this issue, Federated Learning (FL) \citep{yang2019federated,yang2019federated_acm,kairouz2021advances} has been proposed to learn a powerful model without sharing private original data among clients. 
In general, most prior FL works often require frequent model communications to exchange models between local clients and a global server, resulting in heavy communications burden caused by the decentralized data \citep{wu2021node,Chencheng2022_iclr2022}. 
Therefore, it is highly desirable to obtain a powerful federated model with only a few communication rounds.


\begin{figure}[htb]
\centering
\includegraphics[width=1.\linewidth]{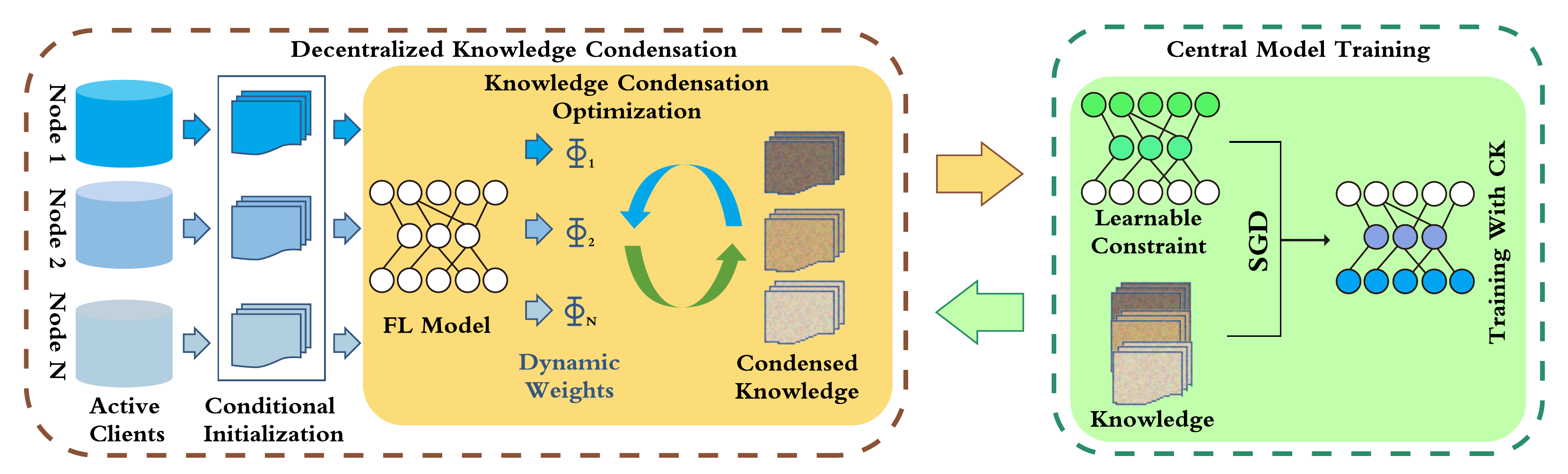}
\caption{Illustration of our pipeline, in which only three active clients are shown. In our method, the local clients conduct meta knowledge condensation from local private data, and the server utilizes the uploaded meta knowledge for training a global model. The local meta knowledge condensation and central model training are conducted in an iterative manner.  For meta knowledge extraction on clients, we design two mechanisms, \textit{i.e.}, meta knowledge sharing, and dynamic weight assignment. For server-side central model training, we introduce a learnable constraint.
}
\label{fig:compa}
\end{figure}


In this work, we propose a new meta knowledge-driven federated learning paradigm to achieve an effective yet communication-efficient model, thus significantly reducing communication costs.
Unlike prior works, we formulate federated learning in a new perspective, where representative information will be distilled from original data and sent to the server for model training.
On the client side, we extract representative information of original data and condense it into a tiny set of highly-compressed synthetic data, namely \textit{meta knowledge}. 
Furthermore, we develop two mechanisms, \textit{i.e.}, dynamic weight assignment and meta knowledge sharing, in the condensation process to mitigate the data heterogeneity issue widely existing in decentralized data.
On the server side, we train our global model with meta knowledge uploaded from clients rather than simply averaging client models.

Specifically, we firstly distill the task-specific knowledge from private data on local clients and condense it as meta knowledge. 
The meta knowledge condensation process is modeled as a bi-level optimization procedure: the inner-loop minimizes the training loss to update a model; and the outer-loop minimizes the training loss to update meta knowledge. 
In the optimization process, we assign dynamic weights to each sample based on its training loss. 
By dynamically adjusting the weight of each sample in training, we empower each sample to contribute adaptively to the current model.
Besides, to further mitigate heterogeneous data distributions among different clients, we design a meta knowledge sharing mechanism. 
The meta knowledge sharing mechanism provides \textit{conditional} initialization to our meta knowledge optimization procedure, thus promoting knowledge exchange among clients.
Since meta knowledge from various clients has been uploaded to the server, our model can be trained with such knowledge that better describe the overall distribution. In contrast, previous methods that average local models on the server do not have such a merit. 
Additionally, we impose a learnable conditional generator on the central model training to improve the training stability. 
Specifically, the generator models the statistical distribution of the uploaded meta knowledge. 
We generate synthetic samples by the modeled distribution, providing historical information to model update.
Note that, as meta knowledge contains the essential information of the original data as well as the corresponding class information, it can be used as normal training data for model training. 
Then, our global model is trained with the \textit{uploaded} and \textit{generated} meta knowledge on the server side.
Consequently, we significantly reduce the impacts of data heterogeneity and decrease the communication rounds.

We conduct extensive experiments on various datasets, including MNIST \citep{lecun2010mnist}, SVHN \citep{netzer2011reading}, CIFAR10 \citep{krizhevsky2009learning}, and CIFAR100~\citep{krizhevsky2009learning}.
The experimental results demonstrate the effectiveness and efficiency of our proposed method. 
Specifically, our method outperforms competing works by a remarkable margin on all the datasets, especially when facing limited communication budgets (\ie, $10$ communication rounds).
Overall, our key contributions are summarized as follows:
\begin{itemize}
    \item We propose a new meta knowledge driven federated learning paradigm. We transmit highly condensed meta knowledge extracted from various local clients to the global server for model training, thus significantly mitigating data heterogeneity issues in federated learning.
    \item We design a federated meta knowledge extraction method that can effectively encode local data for global model training. Moreover, we propose a dynamic weight assignment scheme to promote informativeness of meta knowledge, and a knowledge sharing strategy to exchange meta knowledge among clients without leaking original data. 
    \item We introduce a server-side conditional generator that models the statistical distribution of uploaded meta knowledge to stabilize the training process. Benefiting from the extracted meta knowledge and learned statistical distribution, our model requires fewer communication rounds compared to competing methods while achieving superior performance.
\end{itemize}
\section{Methodology}

\subsection{Problem Definition}
\label{subsec:notations}

Federated Learning (FL) trains a model across a set of decentralized devices, \textit{i.e.}, a set of clients and a server.
Suppose there is data $\mathcal{D}$ distributed on $C$ clients, each of which has a local private training dataset $\mathcal{D}^{c} = \{x^{c}_{i}, y^{c}_{i}\}$, $1 \leq i \leq n^{c}$ and a weight value $p^{c}$. 
It is noted that $\mathcal{D}= \cup \mathcal{D}^{c}$ and $\sum\limits_{c=1}^{C} p^{c}=1$.  Without loss of generality, we discuss a multi-class classification problem under a federated learning setting. 
The learning target is formulated as follows:
\begin{equation}\label{eq:fl_vanilla}
\min\limits_{\textbf{w}}\{\mathcal{L(\textbf{w}, \mathcal{D})} \triangleq \sum \limits_{c=1}^{C} p^{c} \mathcal{L}^{c} (\textbf{w}, \mathcal{D}^{c}) \} ,
\end{equation}
where $\mathcal{L}^{c} (\cdot)$ is a local objective function that is optimized on the $c$-${th}$ client. The loss function $\mathcal{L}^{c} (\cdot)$  is formulated as follows:
\begin{equation}\label{eq:l_formulation}
\mathcal{L}^{c} (\textbf{w}, \mathcal{D}^{c}) \triangleq\frac{1}{n^c} \sum \limits_{i=1}^{n^c} \ell (\textbf{w}, x^{c}_{i}, y^{c}_{i}) ,
\end{equation}
where $\ell(\cdot)$ is a user-defined loss function (\textit{e.g.}, cross-entropy loss), and $\textbf{w}$ denotes model parameters. As most FL algorithms need to exchange locally trained models multiple times, communication burden is often non-negligible. Though one-shot FL \citep{zhou2020distilled_arxiv} has been proposed to reduce communication cost, it suffers performance degradation.

\subsection{Federated Learning via Meta Knowledge}
\label{subsec:ourmethod}
We propose a new FL paradigm to solve the aforementioned limitations. 
Specially, as shown in Figure \ref{fig:compa}, our method conducts federated meta knowledge extraction (FMKE) on local clients and server-side central model training (CMT). 
To mitigate the data heterogeneity issue in FMKE, we design two mechanisms, \textit{i.e.}, dynamic weight assignment, and meta knowledge sharing.
To stabilize the central model training, we introduce a learnable constraint modeled by a conditional generator. 
The technical details are provided in the following subsections.

\subsubsection{Federated Meta Knowledge Extraction on Clients}
\label{subsec:FMKE}
We design Federated Meta Knowledge Extraction (FMKE) to extract key information from decentralized data.
In the decentralized scenario, the original data $\mathcal{D}$ is distributed on a set of clients.
Each client has its private data $\mathcal{D}^{c}$ ($\mathcal{D} = \cup \mathcal{D}^{c}$, $1 \leq c \leq C$, where $C$ is the client number), and a model downloaded from a server. 
For simplifying the following discussion, we denote the downloaded model on the client $c$ as  $\textbf{w}^{c}$. 

On each active local client $c$, FKME distills key information from corresponding  local private data $\mathcal{D}^{c}$. 
The distilled information is condensed as meta knowledge $\hat{\mathcal{D}}^{c}$, which will be used to replace the original data $\mathcal{D}^{c}$ in global model training  \textit{on the server \footnote{The dimension of synthesized $\hat{\mathcal{D}}^{c}_{i}$ is the same as that of original data ${\mathcal{D}}^{c}_{i}$.}}. 
The cardinality of $ \hat{\mathcal{D}}^{}$, \textit{i.e.}, the size of extracted meta knowledge from all active clients, is much less than that of $ \mathcal{D}^{}$.
The condensed meta knowledge $\hat{\mathcal{D}}$ is highly compressed and representative.
 
To learn the meta knowledge $\hat{\mathcal{D}}^{c}$ on active local client $c$, a bi-level optimization solution \citep{rajeswaran2019meta_nips2019,wang2018dataset_arxiv2018} is employed. 
To be specific, the meta knowledge extraction is formulated as a nested optimization process: in the inner-loop, based on an initialized meta knowledge, a model is updated to minimize the training loss over the meta knowledge; 
in the outer-loop, given the updated model, the meta knowledge is renewed by minimizing the training loss over the original data. 
The optimization process on client $c$ is formulated as follows:
\begin{equation}
\hat{\mathcal{D}}^{c}= \argmin_{\mathcal{{D}}^{c} } \mathcal{L}^{c}(\textbf{w}^{*}, \mathcal{D}^{c})  \quad s.t. \quad \textbf{w}^{*}=\argmin_{\textbf{w}^{c}} \mathcal{L}^{c} (\textbf{w}^{c}, \hat{\mathcal{D}^{c}}),
\end{equation}
where $\mathcal{L}^{c}(\cdot,\mathcal{D}^{c})$ denotes a loss function over the original data $\mathcal{D}^{c}$ on client $c$, $\mathcal{L}^{c}(\cdot,\mathcal{\hat{D}}^{c})$ denotes a loss function over the meta knowledge $\hat{\mathcal{D}^{c}}$ on client $c$. 

The inner-loop and outer-loop are implemented alternatingly and stochastic gradient descent (SGD) is employed to update the model and meta knowledge. 
At first, on an active client $c$, we update the model parameter by the following formulation:
\begin{equation}\label{eq:s1_fdd}
\textbf{w}^{c} \gets \textbf{w}^{c} - \eta \bigtriangledown_{} \mathcal{L}^{c} (\textbf{w}^{c}, \hat{\mathcal{D}}^{c}),
\end{equation}  
where $\eta$ denotes a learning rate for the inner-loop.
An updated model $\textbf{w}^{*}$ is obtained in this inner-loop.


Secondly, given the updated model $\textbf{w}^{*}$, we evaluate it on original data $\mathcal{D}^{c}$ and calculate the loss $\mathcal{{L}}^{c}(\textbf{w}^{*}, \mathcal{D}^{c}) $. 
Then the condensed meta knowledge $\hat{\mathcal{D}}^{c}$ can be updated by:
\begin{equation}
\label{eq:s2_fdd}
\hat{\mathcal{D}}^{c} \gets \hat{\mathcal{D}}^{c} - \alpha \bigtriangledown_{} \mathcal{L}^{c}(\textbf{w}^{*}, \mathcal{D}^{c}) ,
\end{equation}
where $\alpha$ denotes a learning rate for the outer-loop, and the meta knowledge $ \hat{\mathcal{D}}^{c}$ is initialized  based on a uniform distribution (\textit{i.e.}, $\hat{\mathcal{D}}^{c}_{ini} \sim U[-1, +1]$).
An updated meta knowledge $\hat{\mathcal{D}}^{c}$ is obtained in this outer-loop. 
The inner-loop and outer-loop are conducted in an alternatingly manner.

Due to the data heterogeneity issue widely existing in FL, the extracted meta knowledge on each client might be biased. 
To mitigate the heterogeneity issue, we design two effective mechanisms in FMKE, namely Dynamic Weight Assignment, and  Meta Knowledge Sharing.

\textbf{Dynamic Weight Assignment}: Concretely, we dynamically assign weights for each training sample in $\mathcal{D}$.
As training progresses, the prediction confidence of each sample varies, making the samples contribute dynamically to the training.
Thus, it is beneficial to treat those samples in an adaptive manner, \textit{i.e.}, based on their prediction output, empowering the samples with different confidence to make different contributions in the current training.
Specifically, we calculate a dynamic weight value $\phi_{i}^{c}$ for each sample $\mathcal{D}_{i}^{c}$, \textit{e.g.}, $ ( x_i^{c}, y_{i}^{c})$, based on its prediction loss $\ell(\textbf{w}^{c}, {x}_{i}^{c}, {y}_i^c)$. 
The formulation is defined as:
\begin{equation}
    \phi_{i}^{c} = \frac{1}{1+ \exp{ (-\tau * \ell(\textbf{w}^{c}, {x}_{i}^{c}, {y}_i^c)}},
\end{equation}
 where $\tau$ is a hyper-parameter to smooth the result,  $\mathcal{D}^{c}=\{\mathcal{D}_{i}^{c}\}, 1 \leq i \leq N^{c}$, and $N^{c}$ denotes original data number on client $c$.
Apparently, the weight of each sample is inversely proportional to its prediction loss in the current model. 
We assign the weight to each sample to update meta knowledge: 
\begin{equation}\label{eq:s2_fdd_dynamic}
    \hat{\mathcal{D}}^{c} \gets \hat{\mathcal{D}}^{c} - \alpha \bigtriangledown_{} \mathcal{L}^{c}(\textbf{w}^{c}, \Phi^{c} \circ \mathcal{D}^{c}),
\end{equation}
where $\mathcal{L}^{c} (\textbf{w}^{c}, \Phi^{c}, \mathcal{D}^{c}) \triangleq\frac{1}{N^c} \sum \limits_{i=1}^{N^c} \phi_{i}^{c} \cdot \ell (\textbf{w}^{c}, x^{c}_{i}, y^{c}_{i}) $, and $\circ$ indicates the weights have been assigned to the corresponding samples.


%



\textbf{Meta Knowledge Sharing}: 
As an optimization process, an appropriate initialization for $\mathcal{\hat{D}}^{c}$ in Eq. \ref{eq:s1_fdd} is crucial to obtain a good final result.
In prior bi-level optimization works \citep{rajeswaran2019meta_nips2019,wang2018dataset_arxiv2018}, the initialization value is randomly sampled from a constant distribution (\textit{i.e.}, $\hat{\mathcal{D}}^{c}_{ini} \sim U[-1, +1]$), namely unconditional initialization.
Due to the heterogeneity issue in FL, the extracted meta knowledge might become biased to the corresponding local data. 
To solve those limitations, we implement a \textit{conditional} initialization in Eq. \ref{eq:s2_fdd_dynamic} with a meta knowledge sharing mechanism.
Conditional initialization \citep{Wang2020_nips2020,Denevi2020_nips2020} requires the initialization value is obtained from data characteristics rather than random generation. 
To achieve conditional initialization, we design a simple yet effective strategy in extracting meta knowledge $\hat{\mathcal{D}}^{c}$ for client $c$ at the current round $t$. 
Concretely, in the process of initializing client $c$, we \textit{randomly} select another client ${c'}$ and use its meta knowledge $\hat{\mathcal{D}}^{c'}_{t-1}$ extracted in the previous round $t-1$ as an initial value in Eq. \ref{eq:s2_fdd_dynamic}.
Correspondingly, the initialization for $\mathcal{\hat{D}}_{}^{c}$ changes from an unconditional manner to a conditional manner: $ {\mathcal{\hat{D}}}_{ini}^{c} \leftarrow \hat{\mathcal{D}}^{c'}_{t-1}, c' \sim randint[1, C], c'\neq c$.
In this manner, the meta knowledge condensation for client $c$ is mainly determined by the local data on client $c$ as well as the knowledge extracted on another client $c'$, mitigating the heterogeneity issue significantly.

 \subsubsection{Server-side Central Model Training}
 \label{subsec:CMT}
After conducting FMKE, we upload the condensed meta knowledge $\mathcal{\hat{D}}$ from clients to a server. 
On the server, the uploaded meta knowledge is used as normal training data to train a {global model $\textbf{W}_{{G}}$}:
\begin{equation}\label{eq:global_training_term1}
\mathcal{L}(\textbf{W}_{G}, \hat{\mathcal{D}})=\frac{1}{|\hat{\mathcal{D}}|}\sum_{\hat{x}_{i},\hat{y}_{i} \in \hat{\mathcal{D}}} \ell( \textbf{W}_{G}, \hat{x}_{i}, \hat{y}_{i}),
\end{equation}
where $\ell(.)$ is a cross-entropy loss function as in Eq.~\ref{eq:l_formulation}, and {$\mathcal{\hat{D}}=\cup \mathcal{\hat{D}}^{c}, 1 \leq c \leq C$}.

To further ameliorate data biases among diverse clients, we introduce additional synthetic training samples into the central model training.
Those introduced training samples are from the same distribution of upload meta knowledge $\mathcal{\hat{D}}$.
Specifically, at first, we model the statistical distribution of uploaded meta knowledge $\hat{\mathcal{D}}$ via a conditional generator, and then we sample additional data points based on the learned distribution. 
Thus, sampled data would share the same distribution as $\hat{\mathcal{D}}$.
After the introduction of sampled synthetic data, we not only stabilize our training procedure but also achieve better performance.

To facilitate the discussion, we divide the model $\textbf{W}_G$ into a feature extractor $\mathcal{F}$ with parameter $\textbf{W}_{G}^{\mathcal{F}}$ and a classifier $\mathcal{C}$ with parameter  $\textbf{W}_{G}^{\mathcal{C}}$, in which $\textbf{W}_{G} = (\textbf{W}_{G}^{\mathcal{F}}, \textbf{W}_{G}^{\mathcal{C}})$. 
Accordingly, we denote a latent representation as $z = \mathcal{F}(\textbf{W}_{G}^{\mathcal{F}}, x)$ and a final prediction as $\hat{y}=\mathcal{C}(\textbf{W}_{G}^{\mathcal{C}}, z)$. 
The conditional generator $\mathcal{G}$ maps a label $y$ into a latent representation $z \sim \mathcal{G}_{}(y, \textbf{w}^{\mathcal{G}})$, and
$\mathcal{G}$ is optimized by the objective:
\begin{equation}
\label{eq:global_training_termG2}
\mathcal{G}^{*}=\argmax_{\mathcal{G}: y \rightarrow z } \mathbb{E}_{y \sim p(y)} \mathbb{E}_{z \sim \mathcal{G}(y, \textbf{w}^{\mathcal{G}}))} \log p(y | z, \textbf{W}_{G}^{\mathcal{C}}),
\end{equation}
where $\textbf{w}^{\mathcal{G}}$ denotes the parameter of $\mathcal{G}$.

The trained generator $\mathcal{G}$ models the distribution of uploaded meta knowledge $\mathcal{\hat{D}}$.
By sampling data from the distribution, we obtain a set of "pseudo" meta knowledge $\mathcal{\hat{D}}^{pseu}$ with corresponding labels.
The generated "pseudo" meta knowledge $\mathcal{\hat{D}}^{pseu}$ as well as  uploaded $\mathcal{\hat{D}}$ are utilized to train the global model by minimizing the following function:
\begin{equation}\label{eq:global_training_term3}
\mathcal{L}_{overall}(\textbf{W}_{G}, \{ \hat{\mathcal{D}}, \hat{\mathcal{D}}^{pseu} \}) = \mathcal{L}(\textbf{W}_{G}, \hat{\mathcal{D}}) + \beta  \mathcal{L}(\textbf{W}_{G}, \hat{\mathcal{D}}^{pseu}),
\end{equation}
where $\beta$ is a parameter and determined by the cardinality fraction $\frac{|\mathcal{\hat{D}}^{pseu}|}{|\mathcal{\hat{D}}|}$.


\paragraph{\textbf{Iterative Symbiosis Paradigm:}}
After central model training, we broadcast the obtained global $\textbf{W}_{G}^{}$ and meta knowledge $\mathcal{\hat{D}}$  to clients. 
On each active client, the broadcasted model $\textbf{W}_{G}$ as well as meta knowledge $\mathcal{\hat{D}}$ are used for a new round of FMKE.  
FMKE and CMT collaborate with each other in an iterative symbiosis paradigm, benefiting each other increasingly as the learning continues.
The pseudo code for our algorithm can be found in the supplementary material.

\paragraph{\textbf{Computational Complexity:}} Our method includes two major parts: federated meta knowledge extraction on clients and global model training on the server. On clients, our method adopts a bi-level optimization to extract the meta-knowledge. The bi-level optimization has a running-time complexity of $O(N \times n)$ \citep{fallah2020convergence_aistats2020}, in which $n$ denotes the meta knowledge size, $N$ denotes the number of samples on the client. On the server, the global model training in our method has a running-time complexity of $O(n)$. In total, the overall running-time complexity of our methods is $O(N \times n)$. 

\section{Experiments}
\subsection{Datasets}
We evaluate our algorithm and compare to the key related works on four benchmarks: MNIST \citep{lecun2010mnist}, SVHN \citep{netzer2011reading}, CIFAR10 \citep{krizhevsky2009learning}, and CIFAR100 \citep{krizhevsky2009learning}.
MNIST is a database of handwritten digits (0-9). 
In MNIST, there are $50,000$ images in the training set and $10,000$ images in the test set, and their size is $1 \times 28 \times 28$ pixels ($c \times w \times h$). 
There are $10$ classes in MNIST dataset. 
SVHN is a real-world image dataset, in which there are $600,000$ color images collected from house numbers in Google Street View images. 
In SVHN, each image is of size $3 \times 32 \times 32$. 
The class number of SVHN is as the same as MNIST. CIFAR10 dataset consists of $60,000$ color images, each of which has a size of $3 \times 32 \times 32$. There are $50,000$ images in the training set and $10,000$ images in the testing set. There are $10$ classes in CIFAR10. CIFAR100 dataset consists of $60,000$ color images, each of which has a size of $3 \times 32 \times 32$. There are $50,000$ images in the training set and $10,000$ images in the testing set. For each image in CIFAR100, there are two kinds of labels, \textit{i.e.}, fine label and coarse label. We choose coarse labels and therefore we have $20$ classes in the experiment on CIFAR100. 

\begin{table}[t]
\centering
\scriptsize
\caption{Results with $10$ rounds.}
\resizebox{0.7\columnwidth}{!}{
\begin{tabular}{lcccccc }
\hline
\textbf{Setting}  & \textbf{FedAvg}  & \textbf{FedProx} & \textbf{FedDistill } &\textbf{FedEnsem}  & \textbf{FedGen} &  \textbf{FedMK}\\
\hline
\multicolumn{7}{c}{\textbf{MNIST}} \\
\hline
$\alpha$=0.50  &74.61\%  & 73.56\%   & 75.04\% & {75.39}\%   & {74.07}\%   & \textbf{92.95}\%\\
$\alpha$=0.75  &73.49\%  & 73.13\%   & 76.21\% & {74.28}\%   & {74.57}\% &  \textbf{92.86}\%\\
$\alpha$=1.0  &74.10\%  & 73.35\%   & 76.19\%  & {74.45}\%   & {73.97}\%   & \textbf{93.63}\%\\
\hline

\multicolumn{7}{c}{\textbf{SVHN}} \\
\hline
$\alpha$=0.50  &29.55\%  & 28.52\%   & 26.92\%   & 29.12\%  & {28.94}\%   & \textbf{74.11}\% \\
$\alpha$=0.75  &31.71\%  & 25.78\%   & 25.77\%  & {29.53}\%  & {30.40}\%   &\textbf{74.90}\%\\
$\alpha$=1.0  &30.87\%  & 29.59\%   & 25.84\%  & {32.67}\% & {33.62}\%   &\textbf{ 74.84}\% \\
\hline

\multicolumn{7}{c}{\textbf{CIFAR10}} \\
\hline
$\alpha$=0.50  &26.63\%  & 26.21\%   & 24.38\%   & 27.56\%  & {25.42}\%   & \textbf{47.33}\% \\
$\alpha$=0.75  &25.42\%  & 24.85\%   & 24.18\%  & 26.42\%  & {26.25}\%   & \textbf{49.04}\%\\
$\alpha$=1.0  &26.80\%  & 26.66\%   & 25.83\%   & 26.74\%  & {25.36}\%  &\textbf{50.32}\%\\
\hline

\multicolumn{7}{c}{\textbf{CIFAR100}} \\
\hline
$\alpha$=0.50  &11.66\%  & 12.09\%   & 10.76\%   & 13.20\%  & {10.34}\%  & \textbf{26.74}\% \\
$\alpha$=0.75  &12.11\%  & 11.65\%   & 11.55\%  & 13.15\%  & {10.21}\%    & \textbf{27.43}\%\\
$\alpha$=1.0  &12.31\%  & 11.34\%   & 11.50\%   & 13.31\%  & {11.19}\%  &\textbf{28.20}\%\\
\bottomrule
\end{tabular}
}
\label{tab:fourdatasets_glob_iter_10}
\vspace{-1em}
\end{table}

\subsection{Implementation Details}
\label{subsec:implementations}


We set the user number to $20$, and the active-user number to $10$. We use $50\%$ of the training set and distribute it on all clients. 
All testing data is utilized for evaluation. 
We use LeNet \citep{lecun1989backpropagation} as the backbone for all methods:{ FedAvg \citep{mcmahan2017communication_aistats2017}, FedProx \citep{lifederated_MlSys2020}, FedDistill \citep{seo2020federated_arxiv2020}, FedEnsemble \citep{zhu2021data_icml2021},  FedGen \citep{zhu2021data_icml2021}, and FedMK. }
Dirichlet distribution $Dir(\alpha)$ is used to model data distributions. 
Specifically, we test three different $\alpha$ values: $0.5$, $0.75$, and $1.0$, respectively. 
We set the communication round number to $10$. 
Learning with a small communication round number (\textit{e.g.}, $10$) denotes learning under a limited communication budget. 
For the methods conducting local model training on clients, \textit{i.e.}, FedAvg, FedProx, FedDistill, FedEnsemble, and FedGen,  we set the local updating number to $20$, and the batch size number to $32$. 
In our method, we set meta knowledge size for each datasets based on their different characteristics (\textit{e.g.}, the class number, sample size, etc): $20$ per class for MNIST; $100$ per class for SVHN; $100$ per class for CIFAR10; $40$ per class for CIFAR100. We run three trials and report the mean accuracy performance (MAP). 


\subsection{Comparative Studies}
\label{subsec:result_comparison}
\textbf{Compared with Prior Works:}
We run all methods under limited communication budgets ($10$ rounds) on four datasets and report the results in Table \ref{tab:fourdatasets_glob_iter_10}. 
As shown in Table \ref{tab:fourdatasets_glob_iter_10}, when communication budgets are limited, all prior works fail to construct a model with a good performance; on the contrary, our method can learn a model outperforming competing works by a remarkable margin. 
For example, on MNIST, our method achieves $93.63\%$ ($\alpha=1$) MAP, outperforming the competing works significantly. 
On SVHN, our method achieves $75.85\%$ ($\alpha=1$) MAP in $10$ rounds, while all competing works have not converged yet.
{We believe that the superior performance of our method comes from those aspects: 1) unlike competing methods conducting local model training with local private data, our FedMK trains a global model based on meta-knowledge from all active clients. Training on knowledge from all active clients makes our method less biased; 2) the designed two mechanisms, \textit{i.e.}, dynamic weight assignment and meta knowledge sharing, make the meta knowledge extraction more stable, leading to a better performance than competing methods \footnote{A more detailed comparison with FedGen can be found in the supplementary document.}.}


\begin{figure}[t]
\centering
\includegraphics[width=1\linewidth,keepaspectratio]{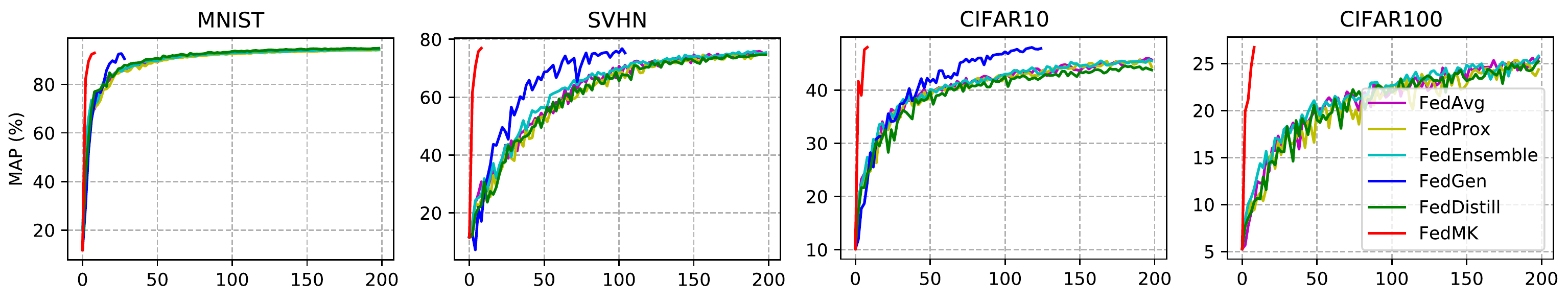}
\caption{Convergence rate comparisons on MNIST, SVHN, CIFAR10, and CIFAR100. $\alpha=1.0$. The x-axis represents communication round numbers.}
\label{fig:convergence_curve}
\end{figure}

\begin{figure}[t]
\centering
\includegraphics[width=1.0\linewidth,keepaspectratio]{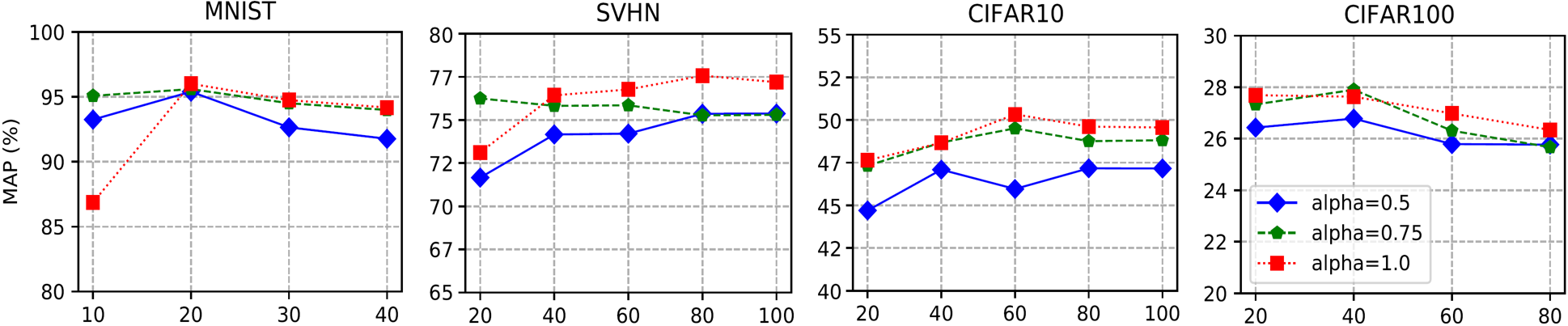}
\caption{Impact of meta knowledge size on final performance. \#round=10. The x-axis represents meta knowledge sizes.}
\label{fig:distilledimagenumberr_mnist}
\vspace{-0.5em}
\end{figure}

\begin{table}[!t]
\centering
\scriptsize
\caption{Impact of each designed mechanism.}
\resizebox{0.85\columnwidth}{!}{
\begin{tabular}{lccccc}
\toprule
 & \textbf{w/o Iter }  & \textbf{w/o Sharing} & \textbf{w/o pseudo knowledge} & \textbf{w/o dynamic weights} &\textbf{Ours}  \\
\midrule
\textbf{CIFAR10}  &28.15\%  & 45.79\%   & 46.71\% & {47.16}\% &  \textbf{47.33}\%  \\
\textbf{CIFAR100}  &18.82\%  & 25.56\%   & 26.10\%  & 26.43\%  & \textbf{26.74}\%  \\
\bottomrule
\end{tabular}
}
\label{tab:ablation}
\vspace{-1.5em}
\end{table}

\label{subsec:ablation}

\textbf{Convergence rate comparisons:}
{We run all methods by $200$ rounds to compare their convergence rates.
We show the performance curves with respect to communication rounds in Figure \ref{fig:convergence_curve}. 
As expected, our method achieves a high convergence speed compared to all competing methods. 
On all datasets, our method achieves satisfactory performance in much fewer communication rounds. 
The results shown in both Table \ref{tab:fourdatasets_glob_iter_10} and Figure \ref{fig:convergence_curve} demonstrate the effectiveness and efficiency of our proposed method.}

\textbf{Impact of the meta knowledge size:} We explore the impact of different meta knowledge sizes on four datasets. 
We draw the performance comparisons in Figure \ref{fig:distilledimagenumberr_mnist}. 
It can be seen that the final performance changes as the meta knowledge size varies. 
For example, on CIFAR10, the MAP score improves as the meta knowledge size increases. 
Another finding is that the optimal meta knowledge sizes between different datasets might be different. 
For example, we achieve the highest MAP score when setting the meta knowledge size to $20$ on MNIST; while on the other three datasets, we obtain the best MAP scores using different meta knowledge sizes.
{More discussions about why meta knowledge decreases the required communication round, and why increasing meta knowledge size will not necessarily improve final performance, can be found in the {supplementary material}. }

\begin{figure}[t]
\centering
\scalebox{1}[1]{\includegraphics[width=1\linewidth]{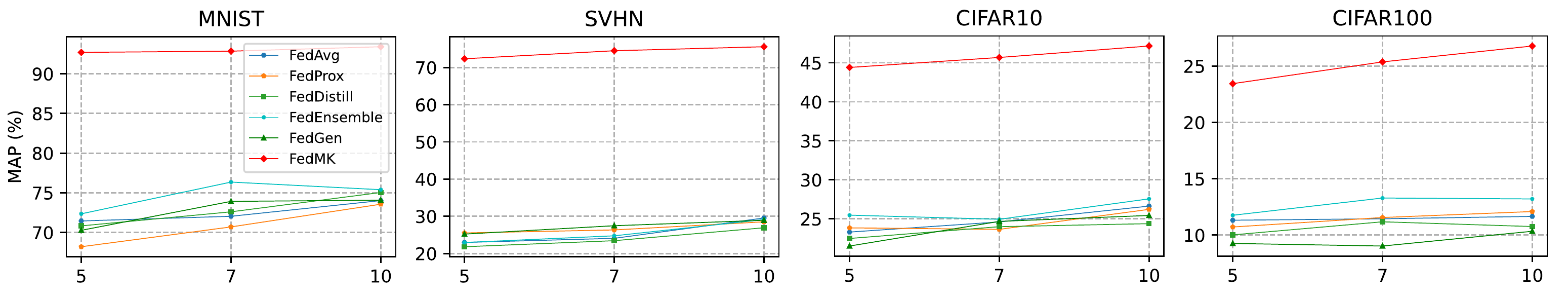}}
\caption{The performance curve w.r.t. the active client number. $\alpha$ = 0.5, \#node  = 20, \#round = 10. The x-axis represents the active client numbers.}
\label{fig:activenode_mnist}
\vspace{-1em}
\end{figure}

\textbf{Impact of active client numbers}: We explore the different active client numbers.
We set the number to $5$, $7$, and $10$, respectively. 
The performance curves are shown in Figure \ref{fig:activenode_mnist}. 
As shown in the figure, our method outperforms all competing works significantly with all different active client numbers. 



\begin{figure}[!t]
\centering
\scalebox{1}[0.9]{\includegraphics[width=1.0\linewidth,keepaspectratio]{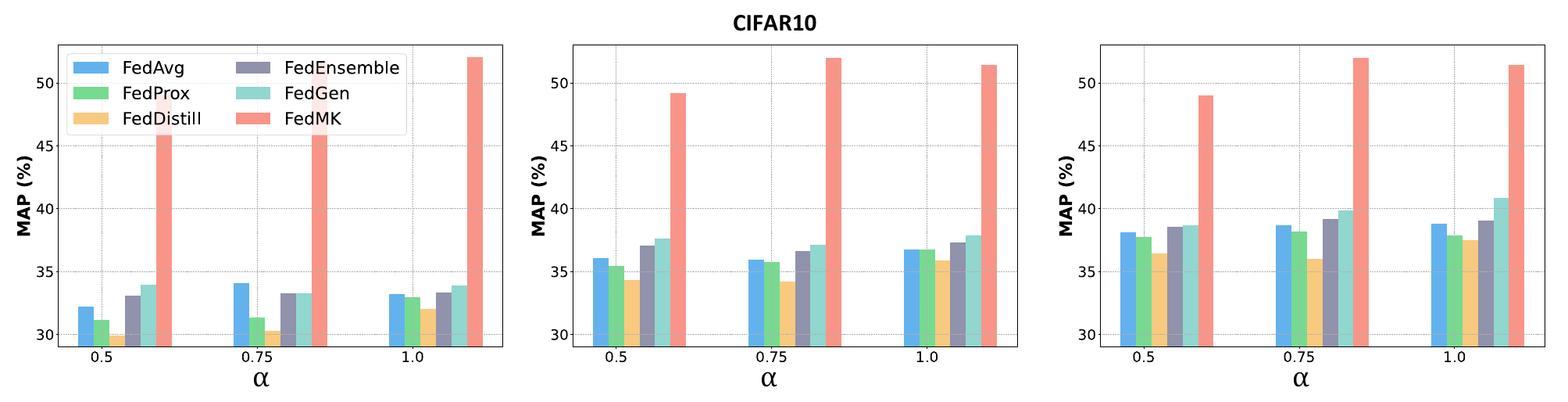}}
\caption{Performance comparisons w.r.t. different round numbers. Left column: $20$ rounds; middle column: $30$ rounds; right column: $40$ rounds. The x-axis represents different $\alpha$ values.}
\label{fig:communication40_20}
\end{figure}

\textbf{Impact of communication rounds:}
We conduct an experiment on CIFAR10 to analyze the impact of communication rounds. Specifically, we set the communication round number to  $20$, $30$, and $ 40$ for comparisons. 
The comparison results are shown in Figure \ref{fig:communication40_20}. 
We can find that our method consistently outperforms competing works by a remarkable margin. 

\textbf{Impact of node numbers:}
{We conduct experiments on SVHN to analyze the impact of node numbers. Specifically, we set the node number to  $200$ for comparisons. Intuitively, when the number of user becomes larger, the number of data on each client will be sparse. And thus, the learning will become difficult, leading to a performance drop. In our experiment, we find that when we increase the node number to $200$, our method ($70.51\%$) still outperforms the FedGen ($66.31\%$) by a remarkable margin.}
\textbf{Impact of designed mechanisms}: We conduct a study on CIFAR10 and CIFAR100 to explore the impact of each designed mechanism. 
We report the performance in Table \ref{tab:ablation}, where "w/o Iter" means that we conduct local meta knowledge condensation and global model update only once \footnote{"w/o Iter" can be regarded as one-shot FL like \cite{zhou2020distilled_arxiv}}; "w/o Sharing" means that we do not adopt meta knowledge sharing between clients; "w/o pseudo meta knowledge" means that there is no learnable constraint on the server; "w/o dynamic weights" means that we update $\hat{\mathcal{D}}$ by treating all samples in $\mathcal{D}$ equally.



As shown in Table \ref{tab:ablation}, there is a drastic performance degradation when there is no iterative collaboration between meta knowledge extraction and central model training ("w/o Iter"). 
Making the meta knowledge condensation and central model training running in an iterative manner improves the performance significantly, \textit{e.g.}, on CIFAR10, MAP score increases from $28.15\%$ to $45.79\%$.  
Moreover, the two designed mechanisms, \textit{i.e.}, the meta knowledge sharing between clients, dynamic weights assignment, significantly boost the MAP score.

\textbf{Evaluation on the pathological non-iid setting:}
{We conduct an experiment on MNIST under the pathological non-iid setting \citep{huang2021personalized_aaai2021}. In the experiment, we set the node number to $20$, the active node number to $10$, the number of classes on each client to $5$, $\alpha$ to $1.0$. We compare all methods under a limited budget communication (\textit{i.e.}, 10 rounds). As seen in Table \ref{tab:pathological}, our method achieves better performance compared to the competing methods.
}

\begin{table}[!t]
\centering
\scriptsize
\caption{Result comparisons on MNIST under the pathological non-iid setting.}
\resizebox{0.8\columnwidth}{!}{
\begin{tabular}{lcccccc}
\toprule
 & \textbf{FedAvg }  & \textbf{FedProx} & \textbf{FedEnsem} & \textbf{FedDistill} & \textbf{FedGen} &\textbf{Ours}  \\
\midrule
\textbf{MAP (\%)} &75.11\%  &72.99\%  & 76.43\%   & 71.19\% & {86.44}\% &  \textbf{88.85}\%  \\
\bottomrule
\end{tabular}
\label{tab:pathological}
}
\end{table}


\textbf{Communication cost:}
We analyze the communication cost of our method on MNIST.
For FedMK, we need to upload and download the meta knowledge, and download a trained model. In the MNIST experiment, the meta image size is $20$ per class, each of which is $4$ bytes $\times$ $28$ $\times$ $28$ (sizeof(float32) $\times$ $w\times h$). 
Therefore, in each communication round, the uploading cost of FedMK is $4$ bytes $\times$ $28$ $\times$ $28$ $\times$ $20$ $\times 10$=$627.2$K (meta knowledge), and the downloading cost is $627.2$K + $105$K $\times 10$= $1677.2$K (meta knowledge and a model). 
In total, the communication cost for FedMK is ($627.2$K+$1,677.2$K) $\times 10$ (\# rounds)=$16$M.
As shown in Figure \ref{fig:convergence_curve}, to achieve comparable performance on MNIST, FedAvg has to run around $200$ rounds. Under this circumstance, the communication cost for FedAvg is $105K \times 10 \times 2 \times 200$  (model size $\times$ \#active node $\times$ $2$ $\times$ \#communication)=420M. 
This indicates that to obtain models with good performance, FedAvg and its variants require much higher communication cost than ours.

\begin{figure}[!t]
\centering
\includegraphics[width=1.0\linewidth,keepaspectratio]{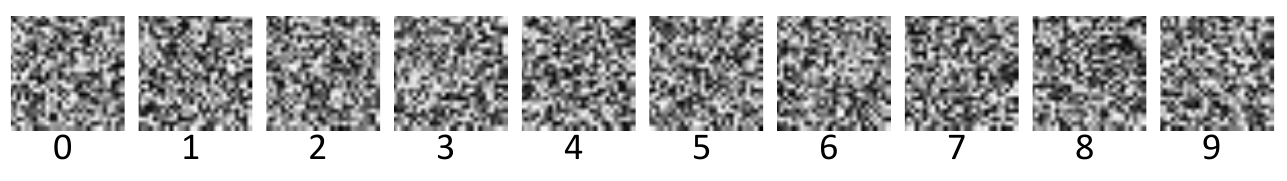}
\caption{The visualization of condensed meta knowledge for MNIST. Each image corresponds to the condensed meta knowledge encoded from images of a specific class in MNIST.}
\label{fig:visual_ck}
\end{figure}

\textbf{Visualization of condensed meta knowledge} We visualize the condensed meta knowledge coded from MNIST dataset. As shown in Figure \ref{fig:visual_ck},
{it is difficult to infer the semantic information from those meta knowledge.
Furthermore, we conduct an experiment based on Deep Leakage \citep{zhu2020deep} to explore the possibility of restoring original data from meta knowledge. We report the results in the {supplementary document}, in which we can find that it is hard to construct correspondence between entries in restored data and original data. 
} 



\section{Related Work}
\textbf{Federated Learning.} As a privacy-preserving solution, federated learning \citep{li2019survey_arxiv2019} provides a new training manner to learn models over a collection of distributed devices. 
In federated learning, the data is located on distributed nodes and not shared.
Learning a model without exchanging local data between nodes minimizes the risk of data leakage but increases the difficulty of training. 
As a representative method, FedAvg \citep{mcmahan2017communication_aistats2017} is proposed to obtain a global model by aggregating local models trained on active clients.
To improve the local training, works such as \citep{lifederated_MlSys2020,zhu2021data_icml2021,li2021model_cvpr2021,Tian2022_arxiv2022} design various strategies in their solutions. All those methods require a large number of communication round to construct a model with satisfactory performance, resulting in heavy communication burden.

Recently, several methods \citep{fallah2020personalized_nips2020,dinh2020personalized_nips2020,seo2020federated_arxiv2020,li2021model_cvpr2021,zhu2021data_icml2021,fan2021federated, gou2021knowledge_ijcv2021,pmlr-v162-dai22b-icml2022,zhang2022fine_cvpr2022} have been proposed to handle the data heterogeneity issue, which can be categorized into personalization-based methods and knowledge distillation-based methods. 
{The personalized FL methods \citep{fallah2020personalized_nips2020,dinh2020personalized_nips2020,pmlr-v162-dai22b-icml2022} aim to learn client-specific models while our goal is to obtain a global model across different clients.}
Knowledge distillation based FL methods \citep{seo2020federated_arxiv2020,zhu2021data_icml2021,zhang2022fine_cvpr2022} propose to share knowledge between nodes, and thus the local training is regularized by local private data and knowledge on other clients.  

\textbf{Compact Data Representation.}
Generally, prior works compressing a large scale data into a small set can be categorized into two main branches: data selection and data compression. 
Data selection methods \citep{rebuffi2017icarl_cvpr2017,castro2018end_eccv2018,aljundi2019gradient_nips2019,sener2017active_iclr2018} select the most representative samples from the original data based on predefined criteria. 
The selected samples are utilized to replace the original data for model training. 
However, how to select an appropriate criterion based on the given data and task is not a trivial issue.
To overcome the aforementioned limitations, synthesizing new samples 
rather than selecting existing samples becomes a more preferable solution. 
\citep{wang2018dataset_arxiv2018,zhao2020dataset_iclr2021, Zhao2021_icml2021,cazenavette2022dataset_cvpr2022} design different solutions for generating synthetic data based on given datasets. 
In those methods, the generated data can replace the original data in the model construction process.
However, those prior synthetic data generation works require data to be localized {in a centralized manner}. It is not trivial to directly apply those methods under a federated learning setting.

\section{Conclusion}
In this paper, we present a new federated learning paradigm driven by meta knowledge, dubbed FedMK, to obtain an effective and fast-converging model. With the help of the proposed paradigm, FedMK can train a powerful model even under a limited communication budget (\textit{e.g.}, 10 communication rounds), decreasing the communication cost significantly. Moreover, our designed mechanisms, \textit{i.e.}, meta knowledge sharing, dynamic weight assignment, and linear constraint, collectively facilitate the central model training,   benefiting FedMK outperforming all competing methods on four datasets. 
\small

\bibliography{iclr2023_conference}

\begin{thebibliography}{41}
\providecommand{\natexlab}[1]{#1}
\providecommand{\url}[1]{\texttt{#1}}
\expandafter\ifx\csname urlstyle\endcsname\relax
  \providecommand{\doi}[1]{doi: #1}\else
  \providecommand{\doi}{doi: \begingroup \urlstyle{rm}\Url}\fi

\bibitem[Aljundi et~al.(2019)Aljundi, Lin, Goujaud, and
  Bengio]{aljundi2019gradient_nips2019}
Rahaf Aljundi, Min Lin, Baptiste Goujaud, and Yoshua Bengio.
\newblock Gradient based sample selection for online continual learning.
\newblock In \emph{NeurIPS}, 2019.

\bibitem[Battiti(1992)]{battiti1992first}
Roberto Battiti.
\newblock First-and second-order methods for learning: between steepest descent
  and newton's method.
\newblock \emph{Neural computation}, 1992.

\bibitem[Castro et~al.(2018)Castro, Mar{\'\i}n-Jim{\'e}nez, Guil, Schmid, and
  Alahari]{castro2018end_eccv2018}
Francisco~M Castro, Manuel~J Mar{\'\i}n-Jim{\'e}nez, Nicol{\'a}s Guil, Cordelia
  Schmid, and Karteek Alahari.
\newblock End-to-end incremental learning.
\newblock In \emph{ECCV}, 2018.

\bibitem[Cazenavette et~al.(2022)Cazenavette, Wang, Torralba, Efros, and
  Zhu]{cazenavette2022dataset_cvpr2022}
George Cazenavette, Tongzhou Wang, Antonio Torralba, Alexei~A Efros, and
  Jun-Yan Zhu.
\newblock Dataset distillation by matching training trajectories.
\newblock In \emph{CVPR}, 2022.

\bibitem[Chencheng et~al.(2022)Chencheng, Zhiwei, Minlie, and
  Tao]{Chencheng2022_iclr2022}
Xu~Chencheng, Hong Zhiwei, Huang Minlie, and Jiang Tao.
\newblock {Acceleration of Federated Learning with Alleviated Forgetting in
  Local Training}.
\newblock In \emph{ICLR}, 2022.

\bibitem[Dai et~al.(2022)Dai, Shen, He, Tian, and
  Tao]{pmlr-v162-dai22b-icml2022}
Rong Dai, Li~Shen, Fengxiang He, Xinmei Tian, and Dacheng Tao.
\newblock {D}is{PFL}: Towards communication-efficient personalized federated
  learning via decentralized sparse training.
\newblock In \emph{ICML}, 2022.

\bibitem[Denevi et~al.(2020)Denevi, Pontil, and Ciliberto]{Denevi2020_nips2020}
Giulia Denevi, Massimiliano Pontil, and Carlo Ciliberto.
\newblock {The Advantage of Conditional Meta-Learning for Biased Regularization
  and Fine Tuning}.
\newblock In \emph{NeurIPS}, 2020.

\bibitem[Dinh et~al.(2020)Dinh, Tran, and
  Nguyen]{dinh2020personalized_nips2020}
Canh~T Dinh, Nguyen~H Tran, and Tuan~Dung Nguyen.
\newblock Personalized federated learning with moreau envelopes.
\newblock In \emph{NeurIPS}, 2020.

\bibitem[Fallah et~al.(2020{\natexlab{a}})Fallah, Mokhtari, and
  Ozdaglar]{fallah2020convergence_aistats2020}
Alireza Fallah, Aryan Mokhtari, and Asuman Ozdaglar.
\newblock On the convergence theory of gradient-based model-agnostic
  meta-learning algorithms.
\newblock In \emph{AISTATS}, 2020{\natexlab{a}}.

\bibitem[Fallah et~al.(2020{\natexlab{b}})Fallah, Mokhtari, and
  Ozdaglar]{fallah2020personalized_nips2020}
Alireza Fallah, Aryan Mokhtari, and Asuman Ozdaglar.
\newblock Personalized federated learning with theoretical guarantees: A
  model-agnostic meta-learning approach.
\newblock In \emph{NeurIPS}, 2020{\natexlab{b}}.

\bibitem[Fan \& Huang(2021)Fan and Huang]{fan2021federated}
Chenyou Fan and Jianwei Huang.
\newblock Federated few-shot learning with adversarial learning.
\newblock In \emph{WiOpt}, 2021.

\bibitem[Gou et~al.(2021)Gou, Yu, Maybank, and Tao]{gou2021knowledge_ijcv2021}
Jianping Gou, Baosheng Yu, Stephen~J Maybank, and Dacheng Tao.
\newblock Knowledge distillation: A survey.
\newblock \emph{International Journal of Computer Vision}, 2021.

\bibitem[Huang et~al.(2021)Huang, Chu, Zhou, Wang, Liu, Pei, and
  Zhang]{huang2021personalized_aaai2021}
Yutao Huang, Lingyang Chu, Zirui Zhou, Lanjun Wang, Jiangchuan Liu, Jian Pei,
  and Yong Zhang.
\newblock Personalized cross-silo federated learning on non-iid data.
\newblock In \emph{AAAI}, 2021.

\bibitem[Kairouz et~al.(2021)Kairouz, McMahan, Avent, Bellet, Bennis, Bhagoji,
  Bonawitz, Charles, Cormode, Cummings, et~al.]{kairouz2021advances}
Peter Kairouz, H~Brendan McMahan, Brendan Avent, Aur{\'e}lien Bellet, Mehdi
  Bennis, Arjun~Nitin Bhagoji, Kallista Bonawitz, Zachary Charles, Graham
  Cormode, Rachel Cummings, et~al.
\newblock Advances and open problems in federated learning.
\newblock \emph{Foundations and Trends{\textregistered} in Machine Learning},
  2021.

\bibitem[Krizhevsky \& Hinton(2009)Krizhevsky and
  Hinton]{krizhevsky2009learning}
Alex Krizhevsky and Geoffrey Hinton.
\newblock Learning multiple layers of features from tiny images.
\newblock 2009.

\bibitem[LeCun et~al.(1989)LeCun, Boser, Denker, Henderson, Howard, Hubbard,
  and Jackel]{lecun1989backpropagation}
Yann LeCun, Bernhard Boser, John~S Denker, Donnie Henderson, Richard~E Howard,
  Wayne Hubbard, and Lawrence~D Jackel.
\newblock Backpropagation applied to handwritten zip code recognition.
\newblock \emph{Neural computation}, 1989.

\bibitem[LeCun et~al.(2010)LeCun, Cortes, and Burges]{lecun2010mnist}
Yann LeCun, Corinna Cortes, and Chris Burges.
\newblock Mnist handwritten digit database, 2010.

\bibitem[Li et~al.(2019)Li, Wen, Wu, Hu, Wang, Li, Liu, and
  He]{li2019survey_arxiv2019}
Qinbin Li, Zeyi Wen, Zhaomin Wu, Sixu Hu, Naibo Wang, Yuan Li, Xu~Liu, and
  Bingsheng He.
\newblock A survey on federated learning systems: vision, hype and reality for
  data privacy and protection.
\newblock \emph{arXiv}, 2019.

\bibitem[Li et~al.(2021)Li, He, and Song]{li2021model_cvpr2021}
Qinbin Li, Bingsheng He, and Dawn Song.
\newblock Model-contrastive federated learning.
\newblock In \emph{CVPR}, 2021.

\bibitem[Li et~al.(2020)Li, Sahu, Zaheer, Sanjabi, Talwalkar, and
  Smith]{lifederated_MlSys2020}
Tian Li, Anit~Kumar Sahu, Manzil Zaheer, Maziar Sanjabi, Ameet Talwalkar, and
  Virginia Smith.
\newblock Federated optimization in heterogeneous networks.
\newblock In \emph{MLSys}, 2020.

\bibitem[McMahan et~al.(2017)McMahan, Moore, Ramage, Hampson, and
  y~Arcas]{mcmahan2017communication_aistats2017}
Brendan McMahan, Eider Moore, Daniel Ramage, Seth Hampson, and Blaise~Aguera
  y~Arcas.
\newblock Communication-efficient learning of deep networks from decentralized
  data.
\newblock In \emph{AISTATS}, 2017.

\bibitem[Netzer et~al.(2011)Netzer, Wang, Coates, Bissacco, Wu, and
  Ng]{netzer2011reading}
Yuval Netzer, Tao Wang, Adam Coates, Alessandro Bissacco, Bo~Wu, and Andrew~Y
  Ng.
\newblock Reading digits in natural images with unsupervised feature learning.
\newblock In \emph{NeurIPS-Workshop}, 2011.

\bibitem[Rajeswaran et~al.(2019)Rajeswaran, Finn, Kakade, and
  Levine]{rajeswaran2019meta_nips2019}
Aravind Rajeswaran, Chelsea Finn, Sham~M Kakade, and Sergey Levine.
\newblock Meta-learning with implicit gradients.
\newblock In \emph{NeurIPS}, 2019.

\bibitem[Rebuffi et~al.(2017)Rebuffi, Kolesnikov, Sperl, and
  Lampert]{rebuffi2017icarl_cvpr2017}
Sylvestre-Alvise Rebuffi, Alexander Kolesnikov, Georg Sperl, and Christoph~H
  Lampert.
\newblock icarl: Incremental classifier and representation learning.
\newblock In \emph{CVPR}, 2017.

\bibitem[Sener \& Savarese(2018)Sener and Savarese]{sener2017active_iclr2018}
Ozan Sener and Silvio Savarese.
\newblock Active learning for convolutional neural networks: A core-set
  approach.
\newblock In \emph{ICLR}, 2018.

\bibitem[Seo et~al.(2020)Seo, Park, Oh, Bennis, and
  Kim]{seo2020federated_arxiv2020}
Hyowoon Seo, Jihong Park, Seungeun Oh, Mehdi Bennis, and Seong-Lyun Kim.
\newblock Federated knowledge distillation.
\newblock \emph{arXiv preprint arXiv:2011.02367}, 2020.

\bibitem[Tian et~al.(2022)Tian, Smith, and Kira]{Tian2022_arxiv2022}
Junjiao Tian, James~Seale Smith, and Zsolt Kira.
\newblock {FedFOR: Stateless Heterogeneous Federated Learning with First-Order
  Regularization}.
\newblock \emph{arXiv}, 2022.

\bibitem[Wang et~al.(2020)Wang, Demiris, and Ciliberto]{Wang2020_nips2020}
Ruohan Wang, Yiannis Demiris, and Carlo Ciliberto.
\newblock {Structured prediction for conditional meta-learning}.
\newblock In \emph{NeurIPS}, 2020.

\bibitem[Wang et~al.(2018)Wang, Zhu, Torralba, and
  Efros]{wang2018dataset_arxiv2018}
Tongzhou Wang, Jun-Yan Zhu, Antonio Torralba, and Alexei~A Efros.
\newblock Dataset distillation.
\newblock \emph{arXiv}, 2018.

\bibitem[Wu \& Wang(2021)Wu and Wang]{wu2021node}
Hongda Wu and Ping Wang.
\newblock Node selection toward faster convergence for federated learning on
  non-iid data.
\newblock \emph{arXiv preprint arXiv:2105.07066}, 2021.

\bibitem[Xie et~al.(2022)Xie, Wang, Zhang, Sato, and
  Sugiyama]{xie2022adaptive_icml2022}
Zeke Xie, Xinrui Wang, Huishuai Zhang, Issei Sato, and Masashi Sugiyama.
\newblock Adaptive inertia: Disentangling the effects of adaptive learning rate
  and momentum.
\newblock In \emph{ICML}, 2022.

\bibitem[Yang et~al.(2019{\natexlab{a}})Yang, Liu, Chen, and
  Tong]{yang2019federated_acm}
Qiang Yang, Yang Liu, Tianjian Chen, and Yongxin Tong.
\newblock Federated machine learning: Concept and applications.
\newblock \emph{ACM Transactions on Intelligent Systems and Technology},
  2019{\natexlab{a}}.

\bibitem[Yang et~al.(2019{\natexlab{b}})Yang, Liu, Cheng, Kang, Chen, and
  Yu]{yang2019federated}
Qiang Yang, Yang Liu, Yong Cheng, Yan Kang, Tianjian Chen, and Han Yu.
\newblock Federated learning.
\newblock \emph{Synthesis Lectures on Artificial Intelligence and Machine
  Learning}, 2019{\natexlab{b}}.

\bibitem[Zhang et~al.(2020)Zhang, Cai, Lin, and
  Shen]{zhang2020deepemd_cvpr2020}
Chi Zhang, Yujun Cai, Guosheng Lin, and Chunhua Shen.
\newblock Deepemd: Few-shot image classification with differentiable earth
  mover's distance and structured classifiers.
\newblock In \emph{CVPR}, 2020.

\bibitem[Zhang et~al.(2022)Zhang, Shen, Ding, Tao, and
  Duan]{zhang2022fine_cvpr2022}
Lin Zhang, Li~Shen, Liang Ding, Dacheng Tao, and Ling-Yu Duan.
\newblock Fine-tuning global model via data-free knowledge distillation for
  non-iid federated learning.
\newblock In \emph{CVPR}, 2022.

\bibitem[Zhao \& Bilen(2021)Zhao and Bilen]{Zhao2021_icml2021}
Bo~Zhao and Hakan Bilen.
\newblock {Dataset Condensation with Differentiable Siamese Augmentation}.
\newblock In \emph{ICML}, 2021.

\bibitem[Zhao et~al.(2021)Zhao, Mopuri, and Bilen]{zhao2020dataset_iclr2021}
Bo~Zhao, Konda~Reddy Mopuri, and Hakan Bilen.
\newblock Dataset condensation with gradient matching.
\newblock In \emph{ICLR}, 2021.

\bibitem[Zhou et~al.(2020)Zhou, Pu, Ma, Li, and Wu]{zhou2020distilled_arxiv}
Yanlin Zhou, George Pu, Xiyao Ma, Xiaolin Li, and Dapeng Wu.
\newblock Distilled one-shot federated learning.
\newblock \emph{arXiv preprint arXiv:2009.07999}, 2020.

\bibitem[Zhou et~al.(2022)Zhou, Nezhadarya, and Ba]{zhou2022dataset}
Yongchao Zhou, Ehsan Nezhadarya, and Jimmy Ba.
\newblock Dataset distillation using neural feature regression.
\newblock \emph{arXiv preprint arXiv:2206.00719}, 2022.

\bibitem[Zhu \& Han(2020)Zhu and Han]{zhu2020deep}
Ligeng Zhu and Song Han.
\newblock Deep leakage from gradients.
\newblock In \emph{NeurIPS}, 2020.

\bibitem[Zhu et~al.(2021)Zhu, Hong, and Zhou]{zhu2021data_icml2021}
Zhuangdi Zhu, Junyuan Hong, and Jiayu Zhou.
\newblock Data-free knowledge distillation for heterogeneous federated
  learning.
\newblock In \emph{ICML}, 2021.

\end{thebibliography}
\bibliographystyle{iclr2023_conference}

\vspace{1cm}

\appendix
\section{Appendix}
\subsection{Additional Experimental Results}
\subsubsection{Datasets}
The additional experiments are also conducted on four benchmarks: MNIST \citep{lecun2010mnist}, SVHN \citep{netzer2011reading}, CIFAR10 \citep{krizhevsky2009learning}, and CIFAR100 \citep{krizhevsky2009learning}.

\subsubsection{Comparative Studies}
\label{subsec:result_comparison}
\textbf{Compared with Prior Works when setting $\alpha$ to $0.1$ and $0.25$:}
We set the $\alpha$ value in Dirichlet distribution $D(\alpha)$ \citep{zhu2021data_icml2021} to $0.1$ and $0.25$, and 
run all methods under limited communication budgets ($10$ rounds) on four datasets. 
We report the results in Table \ref{tab:fourdatasets_glob_iter_10}.
As shown in Table \ref{tab:fourdatasets_glob_iter_10}, when communication budgets are limited ($10$ rounds) and the $\alpha$ value is set to $0.25$ and $0.10$, our method can still learn a model outperforming competing works by a remarkable margin. 

\paragraph{Impact of the smooth parameter $\tau$ in Eq. 6:} We conduct a study on four datasets to explore the impact of smooth parameter $\tau$ in Eq. 6. 
We set the $\tau$ value to $1.0$, $5.0$, and $10.0$, respectively. As shown in Table \ref{tab:fourdatasets_smoothparameter_node10_com10}, we achieve the highest performance in most cases when we set $\tau$ to $5.0$. 
Therefore, we set $\tau$ in Eq. 6 to  $5.0$ for all our experiments.

\subsection{Discussions}

\textbf{Why does utilizing meta-knowledge decrease the required communication rounds?}
{Our method utilizes extracted meta knowledge as normal training data to train a global model on the server. The meta knowledge is extracted from original data via a bi-level optimization, which encodes the "gradient of gradient" with respect to the model. The optimization methods based on the second order gradient generally have a higher convergence speed than the methods using the first order gradient \citep{battiti1992first,xie2022adaptive_icml2022}. Therefore, utilizing meta-knowledge endows our algorithm with a fast convergence speed and decreases the communication round number.}

\textbf{Why increasing the meta knowledge size can not necessarily improve final performance?}
{In the meta knowledge extraction process, the calculated meta-knowledge in each batch represents the average of the model update direction. The average gradient is stable when the batch number increases in a certain range (from ×1 to ×10). As a result, increasing the meta-knowledge sizes does not necessarily increase the performance.
Intuitively, the meta knowledge is highly dense and compressed, encoding the knowledge from original data \citep{zhou2022dataset}. In principle, using the meta knowledge approximates employing the original data. As the amount of information in the original data is constant, the training performance will not necessarily increase as the meta knowledge size increases.}

{ We conduct an experiment on MNIST to show the information change between meta knowledge with different sizes. Concretely, we set the meta-knowledge size (S) as $10$, $20$, $30$, $40$, $50$, $60$, $70$, and $80$, respectively. As earth mover's distance (EMD) has been utilized to compute a structural distance between two data sets to determine their similarity \citep{zhang2020deepemd_cvpr2020}, we use it to evaluate differences according to meta-knowledge with different sizes. The results are listed in Table \ref{tab:knowledgesize}. It can be seen that the EMDs with respect to meta-knowledge sizes are stable, indicating the amount of information in the meta-knowledge does not change significantly with respect to the meta knowledge size.}

\begin{figure}[!t]
\centering
\includegraphics[width=1.0\linewidth,keepaspectratio]{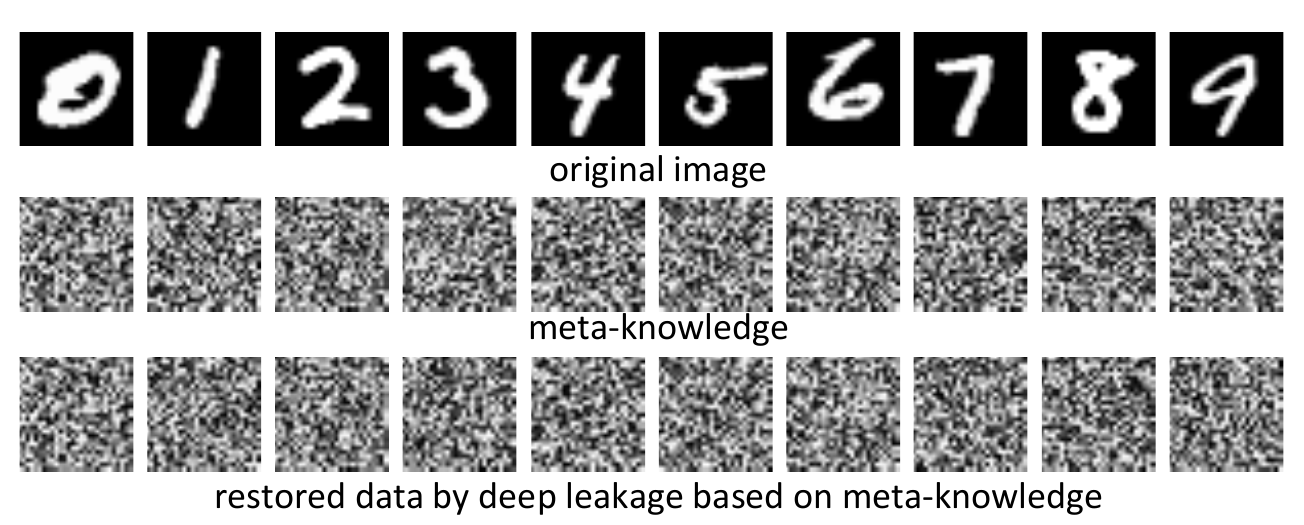}
\caption{The visualization of original images (the top row), extracted meta-knowledge (the middle row), and restored data by deep leakage based on extracted meta-knowledge (the bottom row).}
\label{fig:restored_deepleakage}
\end{figure}

\textbf{The possibility of restoring original data from meta-knowledge.} We conduct an experiment on MNIST to explore the possibility of restoring data from extracted meta-knowledge. The results are shown in Figure \ref{fig:restored_deepleakage}. The original images are shown in the top row, and the extracted meta-knowledge is shown in the middle row. We feed the extracted meta-knowledge and a trained model to Deep Leakage \citep{zhu2020deep}, which is one of the state-of-the-art methods for restoring data from leaked knowledge. The data restored by Deep Leakage is shown in the bottom row. It can be seen that it is hard to construct correspondence between entries in restored data and original data. 


\textbf{The difference between FedGen and FedMK:} There are significant differences between FedGen \citep{zhu2021data_icml2021} and our FedMK, which are listed as follows:

- [Local model training by original data v.s. Meta Knowledge extraction]: FedGen utilizes original data on local clients to train local models, while our method conducts meta knowledge extraction to synthesize meta knowledge, which is used for global model training on a server. In FedGen, the trained local models might diverge due to the data distribution variations among clients.

- [Global model aggregation v.s. Global model training]: FedGen constructs a global model by aggregating uploaded local models; while our method learns a global model based on  meta knowledge uploaded from clients. In our method, the global model learning utilizes knowledge from all active clients, therefore mitigating the bias issue compared to FedGen.

- [The role of conditional generator]: The conditional generator in FedGen is trained on the server and transmitted to clients. On clients, it is used as a constraint in the local model training. On the contrary, the conditional generator in our method is trained and utilized on the server, participating in the global model training. Compared to FedGen, our method has a less communication cost without performance deterioration.
In conclusion, compared to FedGen, our method performs more effectively and efficiently under both practical and pathological non-iid settings.

\begin{table}[!t]\setlength{\tabcolsep}{15pt}
\centering
\scriptsize
\caption{Results with $10$ rounds.}
{
\begin{tabular}{lcccccc }
\hline
\textbf{Setting}  & \textbf{FedAvg}  & \textbf{FedProx} & \textbf{FedDistill } &\textbf{FedEnsem}  & \textbf{FedGen} &  \textbf{FedMK}\\
\hline
\multicolumn{7}{c}{\textbf{MNIST}} \\
\hline
$\alpha$=0.10  &61.95\%  & 61.41\%   & 58.46\%  & {67.89}\%   & {64.83}\% & \textbf{77.37}\%\\
$\alpha$=0.25  &69.52\%  & 68.43\%   & 71.78\%  & {72.23}\%   & {73.41}\% & \textbf{90.07}\%\\
\hline

\multicolumn{7}{c}{\textbf{SVHN}} \\
\hline
$\alpha$=0.10  &20.10\%  & 18.39\%   & 25.44\%  & {24.60}\%   & {24.38}\% & \textbf{57.24}\%\\
$\alpha$=0.25  &23.56\%  & 25.01\%   & 22.70\%  & {23.21}\%   & {28.79}\% & \textbf{65.66}\%\\
\hline

\multicolumn{7}{c}{\textbf{CIFAR10}} \\
\hline
$\alpha$=0.10  &23.71\%  & 21.88\%   & 24.93\%  & {24.80}\%   & {20.16}\% & \textbf{38.45}\%\\
$\alpha$=0.25  &21.85\%  & 22.17\%   & 20.84\%  & {23.98}\%   & {22.94}\% & \textbf{40.75}\%\\
\hline

\multicolumn{7}{c}{\textbf{CIFAR100}} \\
\hline
$\alpha$=0.10  &10.19\%  & 9.41\%   & 12.41\%  & {10.64}\%   & {10.79}\% & \textbf{18.62}\%\\
$\alpha$=0.25  &11.73\%  & 10.43\%   & 8.73\%  & {12.42}\%   & {8.22}\% & \textbf{22.14}\%\\
\hline
\end{tabular}
}
\label{tab:fourdatasets_glob_iter_10}
\end{table}

\begin{table}[t]\setlength{\tabcolsep}{35pt}
\centering
\scriptsize
\caption{Impact of the smooth parameter $\tau$.}
{
\begin{tabular}{lccc }
\hline
  & \textbf{$\tau$}=\textbf{$1.0$}  & \textbf{$\tau$}=\textbf{$5.0$} & \textbf{$\tau$}=\textbf{$10.0$ }\\
\hline
\multicolumn{4}{c}{\textbf{MNIST}} \\
\hline
$\alpha$=0.50  &91.70\%  & \textbf{92.95}\%  & 91.79\% \\
$\alpha$=0.75  &91.90\%  & \textbf{92.86}\%  & 92.23\% \\
$\alpha$=1.0  &91.53\%  & \textbf{93.63}\%  & 91.91\% \\
\hline

\multicolumn{4}{c}{\textbf{SVHN}} \\
\hline
$\alpha$=0.50  &72.24\%  & \textbf{74.11}\%  & 71.60\% \\
$\alpha$=0.75  &71.47\%  & \textbf{74.90}\%  & 74.03\% \\
$\alpha$=1.0  &71.74\%  & \textbf{74.84}\%  & 72.19\% \\
\hline

\multicolumn{4}{c}{\textbf{CIFAR10}} \\
\hline
$\alpha$=0.50  &\textbf{47.72}\%  & {47.33}\%  & 46.62\% \\
$\alpha$=0.75  &48.17\%  & {49.04}\%  & \textbf{49.27}\% \\
$\alpha$=1.0  &47.82\%  & \textbf{50.32}\%  & 48.54\% \\
\hline

\multicolumn{4}{c}{\textbf{CIFAR100}} \\
\hline
$\alpha$=0.50  &{26.06}\%  & \textbf{26.74}\%  & 26.45\% \\
$\alpha$=0.75  &26.93\%  & \textbf{27.43}\%  & 26.98\% \\
$\alpha$=1.0  &25.43\%  & \textbf{28.20}\%  & 26.15\% \\
\hline
\end{tabular}
}
\label{tab:fourdatasets_smoothparameter_node10_com10}
\end{table}

\begin{table}[!t]
\setlength{\tabcolsep}{10pt}
\centering
\scriptsize
\caption{EMDs with respect to meta-knowledge sizes.}
\begin{tabular}{lcccccccc}
\toprule
  \textbf{Meta-Knowledge size(S) }  & \textbf{10} & \textbf{20} & \textbf{30} & \textbf{40} &\textbf{50} &\textbf{60} &\textbf{70} &\textbf{80}  \\
\midrule
\textbf{Difference w.r.t. S=10} &0  &10  & 20   & 30 & 40 & 50 &60 &70  \\
\midrule
\textbf{EMD($meta_{\textbf{S}}, meta_{10})$} &0.000  &0.023  & 0.054   & 0.046 & 0.053 & 0.045 &0.045 &0.043  \\
\bottomrule
\end{tabular}
\label{tab:knowledgesize}
\vspace{-1.5em}
\end{table}

\subsection{Algorithm}
The algorithm of FedMK is illustrated in Alg. \ref{alg:npdg}.

\begin{algorithm}[!t]
  \caption{FedMK }
  \label{alg:npdg}
  \KwIn{Original data $\mathcal{D}$; global parameters $\textbf{W}_{G}$; generator parameter $\textbf{w}^{\mathcal{G}}$; the communication budget .}
  \KwOut{Optimal $\textbf{W}_{G}^{*}$}

\While{not over the communication budget}
  {
    the server selects active clients ${C}$ uniformly at random, broadcasts $\textbf{W}_{G}$ to the selected clients $C$.
    
    $\rhd$Federated Meta Knowledge Extraction on selected clients $C$:
    
    \For{all user $c \in C$ in parallel}
    {
      $\textbf{w}^{c} \leftarrow \textbf{W}_{G}$\;
      
      \For{t = 1, ...,  $\#$Round}
      {
      conduct the conditional initialization: $ {\mathcal{\hat{D}}}_{ini}^{c} \leftarrow \hat{\mathcal{D}}^{c'}_{t-1}, c' \sim randint[1, C], c'\neq c$\;
      
      
     calculate dynamic weights by Eq. 6\;
     generate $\hat{\mathcal{D}}^{c}$ by Eq. 3\;
      }
      
      
      send the $\hat{\mathcal{D}}^{c}$ to the server.
    }
    
          $\rhd$Global Model Training on the server:
      
      update generator parameter $\textbf{w}^{\mathcal{G}}$ by  Eq. 9\;
      generate $\hat{\mathcal{D}}^{pseu}$ by the updated generator ${\mathcal{G}}$\;
      update global parameter $\textbf{W}_{G}$ by  Eq. 10.
      
  }

  return $\textbf{W}_{G}$ as $\textbf{W}_{G}^{*}$\;
\end{algorithm}

\small


\end{document}